\documentclass[10pt,twocolumn,letterpaper]{article}

\usepackage{cvpr}              

\usepackage{graphicx}
\usepackage{amsmath}
\usepackage{amssymb}
\usepackage{booktabs}
\usepackage{bbm}
\usepackage[accsupp]{axessibility}

\usepackage{color, colortbl}
\definecolor{LightGrey}{rgb}{0.88,0.88,0.88}

\usepackage[capitalize]{cleveref}
\crefname{section}{Section}{Sections}
\Crefname{section}{Section}{Sections}
\Crefname{table}{Table}{Tables}
\crefname{table}{Table}{Table}

\def\cvprPaperID{3051} 
\def\confName{CVPR}
\def\confYear{2023}

\begin{document}

\title{Learning Bottleneck Concepts in Image Classification}

\author{Bowen Wang$^{1}$, Liangzhi Li$^{2}\thanks{Corresponding author.}$, Yuta Nakashima$^{2}$, Hajime Nagahara$^{2}$\\
Osaka University, Japan\\
{\tt\small $^1$bowen.wang@is.ids.osaka-u.ac.jp}\\
{\tt\small $^2$\{li, n-yuta, nagahara\}@ids.osaka-u.ac.jp}
}
\maketitle

\begin{abstract}
Interpreting and explaining the behavior of deep neural networks is critical for many tasks. Explainable AI provides a way to address this challenge, mostly by providing per-pixel relevance to the decision. Yet, interpreting such explanations may require expert knowledge. Some recent attempts toward interpretability adopt a concept-based framework, giving a higher-level relationship between some concepts and model decisions. This paper proposes Bottleneck Concept Learner (BotCL), which represents an image solely by the presence/absence of concepts learned through training over the target task without explicit supervision over the concepts. It uses self-supervision and tailored regularizers so that learned concepts can be human-understandable. Using some image classification tasks as our testbed, we demonstrate BotCL's potential to rebuild neural networks for better interpretability \footnote{Code is available at https://github.com/wbw520/BotCL and a simple demo is available at https://botcl.liangzhili.com/.}.
\end{abstract}

\section{Introduction}
Understanding the behavior of deep neural networks (DNNs) is a major challenge in the explainable AI (XAI) community, especially for medical applications \cite{holzinger2019causability,van2022explainable}, for identifying biases in DNNs \cite{wang2019designing,arrieta2020explainable,hirota2022quantifying}, \etc. Tremendous research efforts have been devoted to the post-hoc paradigm for \emph{a posteriori} explanation \cite{selvaraju2017grad,petsiuk2018rise}. This paradigm produces a relevance map to spot regions in the input image that interact with the model's decision. Yet the relevance map only tells \emph{low-level} (or per-pixel) relationships and does not explicitly convey any semantics behind the decision. Interpretation of relevance maps may require expert knowledge. 

The \textit{concept-based} framework \cite{zhou2018interpretable,koh2020concept,stammer2022interactive} is inspired by the human capacity to learn a new concept by (subconsciously) finding finer-grained concepts and reuse them in different ways for better recognition \cite{lake2015human}. Instead of giving per-pixel relevance, this framework offers higher-level relationships between the image and decision mediated by a limited number of \emph{concepts}. That is, the decision is explained by giving a set of concepts found in the image. The interpretation of the decision is thus straightforward once the interpretation of each concept is established.



\begin{figure}[t]
\centering
\includegraphics[width=0.90\columnwidth]{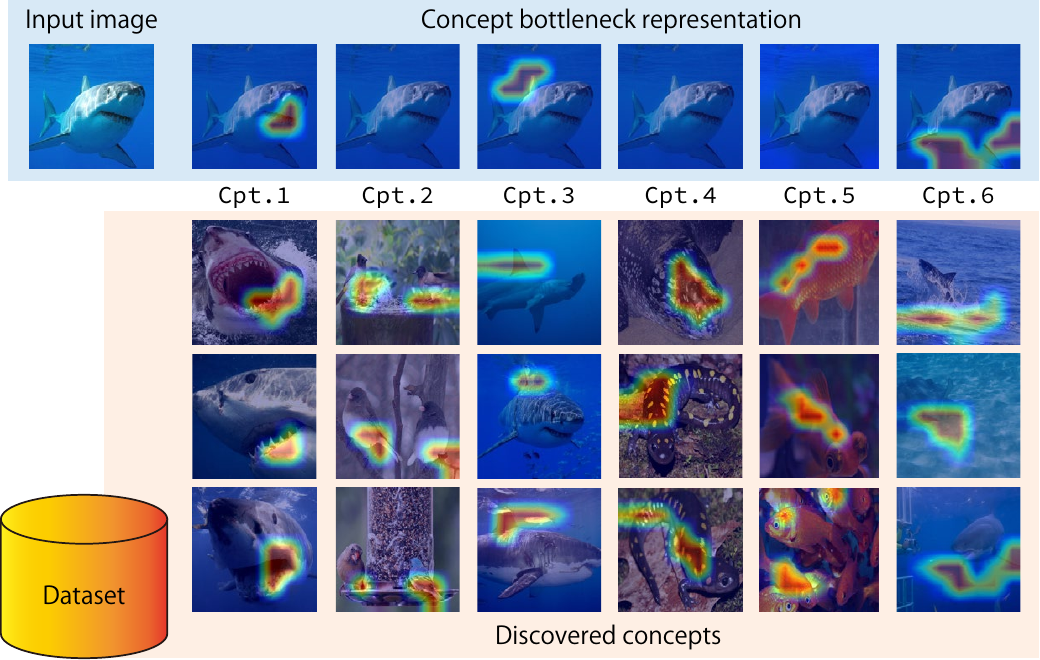}
\caption{Examples of concepts discovered by BotCL in ImageNet \cite{deng2009imagenet}  and concepts in the input image. BotCL automatically discovers a set of concepts optimized for the target task and represents an image solely with the presence/absence of concepts.}
\label{fig:1.1}
\end{figure}

Some works use concepts for the post-hoc paradigm for better interpretation of the decision \cite{zhou2018interpretable,ghorbani2019towards}, while the link between the decision and concepts in the image is not obvious. The \emph{concept bottleneck} structure \cite{kumar2009attribute} uses the presence/absence of concepts as image representation (referred to as \emph{concept activation}).
The classifier has access only to the concept activation, so the decision is strongly tied to the concepts. This bottleneck structure has become the mainstream of the concept-based framework \cite{bucher2018semantic,losch2019interpretability,huang2020interpretable,rigotti2022attention}.  

A major difficulty in this framework is designing a set of concepts that suits the target task. A promising approach is handcrafting them \cite{zhou2014object,bau2017network,kim2018interpretability}, which inherently offers better interpretability at the cost of extra annotations on the concepts. Recent attempts automatically discover concepts \cite{ghorbani2019towards,ge2021peek,alvarez2018towards, NEURIPS2020_ecb287ff}. Such concepts may not always be consistent with how humans (or models) see the world \cite{zhang2018unreasonable,laugel2019dangers} and may require some effort to interpret them, but concept discovery without supervision is a significant advantage. 

Inspired by these works, we propose \textbf{\underline{bot}tleneck \underline{c}oncept \underline{l}earner} (BotCL) for simultaneously discovering concepts and learning the classifier. BotCL optimizes concepts for the given target image classification task without supervision for the concepts. An image is represented solely by the existence of concepts and is classified using them. 
We adopt a slot attention-based mechanism \cite{locatello2020object,li2021scouter} to spot the region in which each concept is found. This gives an extra signal for interpreting the decision since one can easily see what each learned concept represents by collectively showing training images with the detected concepts. Figure \ref{fig:1.1} shows examples from ImageNet \cite{deng2009imagenet}. BotCL discovers a predefined number of concepts in the dataset, which are exemplified by several images with attention maps. An image of \texttt{\small Great White Shark} is represented by the right part of mouth (\texttt{Cpt.1}) and fins (\texttt{Cpt.3}). BotCL uses a single fully-connected (FC) layer as a classifier, which is simple but enough to encode the co-occurrence of each concept and each class. 


\textbf{Contribution}. For better concept discovery, we propose to use \emph{self-supervision over concepts}, inspired by the recent success in representation learning \cite{chen2020simple,he2020momentum}. Our ablation study demonstrates that self-supervision by contrastive loss is the key. We also try several constraints on concepts themselves, \ie, \emph{individual consistency} to make a concept more selective and \emph{mutual distinctiveness} for better coverage of various visual elements. These additional constraints regular the training process and help the model learn concepts of higher quality.


\section{Related Works}\label{sec:relate}

\subsection{Explainable AI}

XAI focuses on uncovering black-box deep neural networks\cite{schulz2020restricting,shrikumar2017learning,wang2020score,bach2015pixel,NEURIPS2020_ecb287ff,fong2019understanding,wang2021mtunet,simonyan2014deep,chattopadhay2018grad}. A major approach is generating a relevance map that spots important regions for the model's decision. Various methods have been designed for specific architectures, \eg, CAM \cite{zhou2016learning}, and Grad-CAM \cite{selvaraju2017grad} for convolutional neural networks;  \cite{chefer2021transformer} for Transformers \cite{vaswani2017attention}. However, the interpretation of the relevance maps may not always be obvious, which spurs different approaches \cite{context,shi2019knowledge}, including context-based ones.

\subsection{Concept-based framework for interpretability}


A straightforward way to define a set of concepts for a target task is to utilize human knowledge \cite{zhou2014object,koh2020concept}. Such concepts allow quantifying their importance for a decision \cite{kim2018interpretability}.  A large corpus of concepts \cite{bau2017network,varshneya2021learning} is beneficial for delving into hidden semantics in DNNs \cite{zhou2018interpretable}. These methods are of the post-hoc XAI paradigm, but a handcrafted set of concepts can also be used as additional supervision for models with the concept bottleneck structure\cite{koh2020concept,rigotti2022attention,he2022transfg}.

Handcrafting a set of concepts offers better interpretability as they suit human perception; however, the annotation cost is non-negligible.  Moreover, such handcrafted concepts may not always be useful for DNNs \cite{zhang2018unreasonable}. These problems have motivated automatic concept discovery. Superpixels are a handy unit for finding low-level semantics, and concepts are defined by clustering them \cite{ghorbani2019towards,ge2021peek,posada2022eclad}. Another interesting approach is designing a set of concepts to be sufficient statistics of original DNN features \cite{NEURIPS2020_ecb287ff}. These methods are designed purely for interpretation, and concept discovery is made aside from training on the target task.

The concept bottleneck structure allows optimizing a set of concepts for the target task. ProtoPNet \cite{chen2019looks} adopts this structure and identifies concepts based on the distance between features and concepts. SENN \cite{alvarez2018towards} uses self-supervision by reconstruction loss for concept discovery. 

SENN inspired us to use self-supervision, but instead of reconstruction loss, we adopt contrastive loss tailored. For a natural image classification task, this contrastive loss is essential for concept discovery.

\begin{figure*}[!t]
  \begin{subfigure}{0pt}\phantomsubcaption\label{fig:st-a}\end{subfigure}
  \begin{subfigure}{0pt}\phantomsubcaption\label{fig:st-b}\end{subfigure}
\centering
\includegraphics[width=1.0\textwidth]{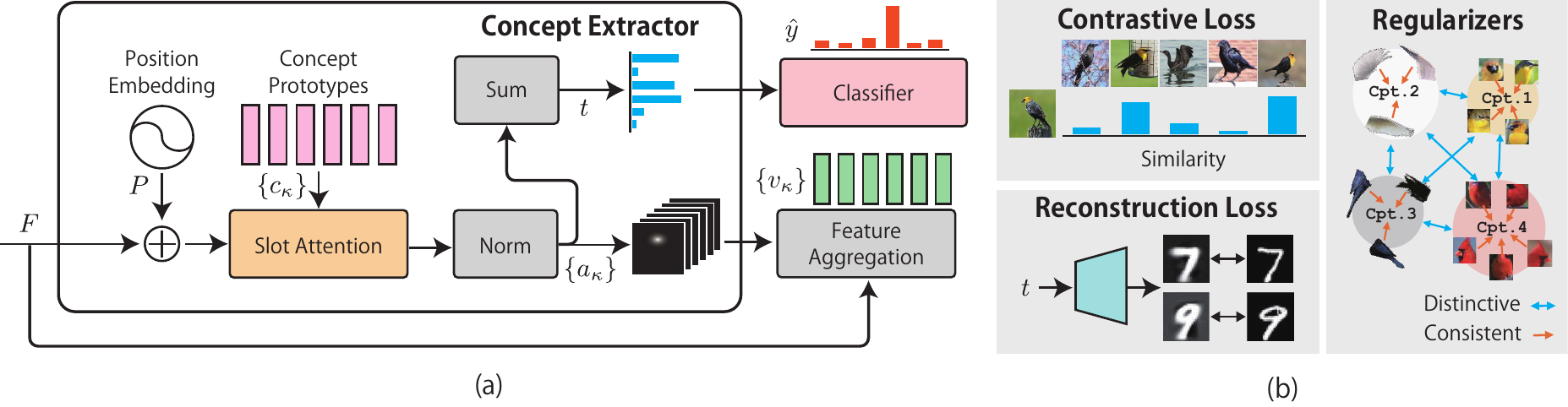}
\caption{(a) The model pipeline. (b) Self-supervision and regularizers.}
\label{fig:structure}
\end{figure*}

\section{Model}

Given a dataset $\mathcal{D} = \{(x_{i}, y_{i}) | i = 1, 2, \dots, N\}$, where $x_i$ is an image and $y_i$ is the target class label in the set $\Omega$ associated with $x_i$. BotCL learns a set of $k$ concepts while learning the original classification task. Figure \ref{fig:st-a} shows an overview of BotCL's training scheme, consisting of a concept extractor, regularizers, and a classifier, as well as self-supervision (contrastive and reconstruction losses). 

For a new image $x$, we extract feature map $F=\Phi(x)\in\mathbb{R}^{d \times h\times w}$ using a backbone convolutional neural network $\Phi$. $F$ is then fed into the concept extractor $g_{C}$, where $C$ is a matrix, each of whose $\kappa$-th column vector $c_\kappa$ is a \emph{concept prototype} to be learned. The concept extractor produces concept bottleneck activations $t \in [0, 1]^k$, indicating the presence of each concept, as well as concept features $V \in \mathbb{R}^{d \times k}$ from regions where each concept exists. The concept activations in $t$ are used as input to the classifier to compute score $s \in [0, 1]^{|\Omega|}$. We use self-supervision and regularizers for training, taking $t$ and $V$ as input to constrain the concept prototypes.

\subsection{Concept Extractor}

Concept extractor uses slot attention \cite{locatello2020object,li2021scouter}-based mechanism to discover visual concepts in $\mathcal{D}$. We first add position embedding $P$ to feature map $F$
to retain the spatial information, 
\ie, $F' = F + P$. The spatial dimension of $F'$ is flattened, so its shape is $l \times d$, where $l = hw$.

The slot-attention computes attention over the spatial dimension for concept $\kappa$ from $c_\kappa$ and $F'$. Let $Q(c_\kappa) \in \mathbb{R}^{d},$ and $K(F') \in \mathbb{R}^{d\times l}$ denote nonlinear transformations for $c_\kappa$ and $F'$, respectively, given as multi-layer perceptrons with three FC layers and a ReLU nonlinearity between them.
Attention $a_\kappa \in [0,1]^{l}$ is given using a normalization function $\phi$ (refer to \textbf{supp. material}) as
\begin{align}
    a_\kappa = \phi(Q(c_\kappa)^\top K(F')).
\end{align}

This attention indicates where concept $\kappa$ presents in the image as shown in Figure \ref{fig:1.1}. If concept $\kappa$ is absent, corresponding entries of $a_\kappa$ are all close to 0.
We summarize the presence of each concept into concept activation $t_\kappa$ by reducing the spatial dimension of $a_\kappa$ as $t_\kappa =  \text{tanh}(\sum_m a_{\kappa m})$,
where $a_{\kappa m}$ is the $m$-th element of $a_\kappa$.

\subsection{Feature Aggregation}
For training, we also aggregate features in $F$ corresponding to concept $\kappa$ into concept feature $v_\kappa$ by   
\begin{equation}
    v_\kappa = Fa_\kappa, 
\end{equation}
which gives the average of image features over the spatial dimension weighted by attention.

\subsection{Classifier}

We use a single FC layer without a bias term as the classifier, and  
concept activation $t = (t_1, \dots, t_k)^\top$ is the only input, serving as the concept bottleneck \cite{koh2020concept}. Formally, letting $W$ be a learnable matrix, prediction $\hat{y} \in \mathbb{R}^{|\Omega|}$ is given by
\begin{equation}
    \hat{y} = W t.
\end{equation}
This classifier can be roughly interpreted as learning the correlation between the class and concepts. Let $w_\omega$ be the raw vector of $W$ corresponding to class $\omega \in \Omega$, and $w_{\omega\kappa}$ is its $\kappa$-th element. A positive value of $w_{\omega\kappa}$ means that concept $\kappa$ co-occurs with class $\omega$ in the dataset, so its presence in a new image positively supports class $\omega$. Meanwhile, a negative value means the concept rarely co-occurs. 

\section{Training}

\subsection{Self-supervision for Concept Discovery}

The absence of concept labels motivates us to incorporate self-supervision for concept discovery. We employ two losses for different types of target tasks. 

\paragraph{Reconstruction loss.}

SENN \cite{alvarez2018towards} uses an autoencoder-like structure for learning better representation. We assume this structure works well when visual elements are strongly tied with the position\footnote{For example, images of ``7'' in MNIST almost always have the acute angle in the top-right part.} since even discrete concepts should have sufficient information to reconstruct the original image. Based on this assumption, we design a reconstruction loss for self-supervision. As shown in Figure \ref{fig:st-b}, decoder $D$ only takes $t$ as input and reconstructs the original image. We define our reconstruction loss as
\begin{equation}
    l_\text{rec} = \frac{1}{|\mathcal{B}|}\sum_{x\in \mathcal{B}} \| D(t) - x\|^{2}.
\end{equation}

\paragraph{Contrastive loss.}

The composition of natural images is rather arbitrary, so information in $t$ should be insufficient to reconstruct the original image. we thus design a simple loss for an alternative, borrowing the idea from the recent success of contrastive learning for self-supervision \cite{chen2020simple,he2020momentum}.

We leverage the image-level labels of the target classification task. Let $\hat{t}=2t-\mathbf{1}_k$, where $\mathbf{1}_k$ is the $k$-dimensional vector with all elements being 1.  If a pair $(\hat{t}, \hat{t}')$ of concept activations belong to the same class (\ie, $y=y'$ for $y$ and $y'$ corresponding to $\hat{t}$ and $\hat{t}'$), they should be similar to each other since a similar set of concepts should be in the corresponding images, and otherwise dissimilar. The number $|\Omega|$ of classes can be smaller than the number $|\mathcal{B}|$ of images in a mini-batch so that a mini-batch can have multiple images of the same class. Therefore, we use sigmoid instead of softmax, leading to
\begin{equation}\label{log_like}
    l_\text{ret} = - \frac{1}{|\mathcal{B}|}\sum \alpha(y, y') \log J(\hat{t},\hat{t}',y,y'),
\end{equation} 
where $\alpha$ is the weight to mitigate the class imbalance problem (see \textbf{supp.~material}) and
\begin{equation}
    J(\hat{t},\hat{t}', y, y') = \begin{cases}
    \sigma(\hat{t}^\top \hat{t}') & \text{for $y = y'$} \\
    1 - \sigma(\hat{t}^\top \hat{t}') & \text{otherwise}    \end{cases}.
\end{equation}

\subsection{Concept Regularizers}\label{regulator}
We also employ concept regularizers to facilitate training. They constrain concept prototypes $\{c_\kappa\}$ through $\{v_\kappa\}$. 

\paragraph{Individual consistency.}

For better interpretability, each learned concept should not have large variations. That is, the concept features $v_\kappa$ and $v'_\kappa$ of different images should be similar to each other if $t_\kappa$ is close to 1. 
Let $\mathcal{H}_\kappa$ denote the set of all concept features of different images in a mini-batch, whose activation is larger than the empirical threshold $\xi$, which is dynamically calculated as the mean of $t_\kappa$ in a mini-batch. Using the cosine similarity $\text{sim}(\cdot , \cdot )$, we define the consistency loss as:
\begin{equation}\label{f1}
    l_\text{con} = - \frac{1}{k}\sum_{\kappa}  \sum_{v_\kappa, v'_\kappa} \frac{\text{sim}(v_{\kappa}, v'_{\kappa})}{|\mathcal{H}_\kappa|(|\mathcal{H}_\kappa|-1)},
\end{equation}
where the second summation is computed over all combinations of concept features $v_\kappa$ and $v'_\kappa$.
This loss penalizes a smaller similarity between $v_\kappa$ and $v'_\kappa$.

\paragraph{Mutual distinctiveness.}
To capture different aspects of images, different concepts should cover different visual elements. This means that the average image features of concept $\kappa$ within a mini-batch, given by $\bar{v}_\kappa = \sum_{v_\kappa \in \mathcal{H}_\kappa} v_\kappa$, should be different from any other $v_{\kappa'}$. We can encode this into a loss term as 
\begin{equation}\label{f2}
    l_\text{dis} = \sum_{\kappa, \kappa'} \frac{\text{sim}(\bar{v}_\kappa, \bar{v}_{\kappa'})}{k(k-1)},
\end{equation}
where the summation is computed over all combinations of concepts. Note that concept $\kappa$ is excluded from this loss if no image in a mini-batch has concept $\kappa$.

\subsection{Quantization Loss}

Concept activation $t$ can be sufficiently represented by a binary value, but we instead use a continuous value for training. We design a quantization loss to guarantee values are close to $0$ or $1$, given by
\begin{equation}
    l_\text{qua} = \frac{1}{k|\mathcal{B}|} \sum_{x \in \mathcal{B}} \left\| \text{abs}(\hat{t}) - \mathbf{1}_\kappa \right\|^2,
\end{equation}
where $\text{abs}(\cdot)$ gives the element-wise absolute value and $\|\cdot\|$ gives the Euclidean norm. 

\subsection{Total Loss}
We use softmax cross-entropy for the target classification task's loss, donated by $l_\text{cls}$. The overall loss of BotCL is defined by combining the losses above as 
\begin{equation}
L =  l_\text{cls} + \lambda_\text{R} l_\text{R} + \lambda_\text{con} l_\text{con} + \lambda_\text{dis} l_\text{dis} + \lambda_\text{qua} l_\text{qua},
\end{equation}
where $l_\text{R}$ is either $l_\text{rec}$ or $l_\text{ret}$ depending on the target domain, $\lambda_\text{qua}$, $\lambda_\text{con}$, $\lambda_\text{dis}$, and $\lambda_{R}$ are weights to balance each term.

\section{Results}
\subsection{Experimental Settings}
We evaluate BotCL on MNIST \cite{deng2012mnist}, CUB200 \cite{welinder2010caltech}, and ImageNet \cite{deng2009imagenet}. For evaluating discovered concepts, we regenerated a synthetic shape dataset (Synthetic) \cite{NEURIPS2020_ecb287ff}. 

For MNIST, we applied the same networks as \cite{alvarez2018towards} for the backbone and the concept decoder. For CUB200 (same data split as \cite{koh2020concept}) and ImageNet, we used pre-trained ResNet \cite{he2016deep} as the backbone with a $1 \times 1$ convolutional layer to reduce the channel number (512 for ResNet-18 and 2048 for ResNet-101) to 128. We chose a concept number $k=20$ for MNIST and $k=50$ for the other natural image datasets. To generate Synthetic, we followed the setting of \cite{NEURIPS2020_ecb287ff}, where 18,000 images were generated for training and 2,000 for evaluation. We used $k=15$ with ResNet-18 backbone.

Images were resized to $256 \times 256$ and cropped to $224 \times 224$ (images in Synthetic were directly resized to $224 \times 224$). Only random horizontal flip was applied as data augmentation during training. The weight of each loss was defaulted to $\lambda_\text{qua} = 0.1$, $\lambda_\text{con}=0.01$, $\lambda_\text{dis}=0.05$, and $\lambda_\text{R} = 0.1$.

\subsection{Classification Performance}
We compare the performance of BotCL with corresponding baselines (LeNet for MNIST and ResNet-18 for others with a linear classifier), our reimplementation of k-means and PCA in \cite{NEURIPS2020_ecb287ff},\footnote{Implementation details are in \textbf{supp.~material.}} and state-of-the-art concept-based models. \cref{acc_tab} summarized the results. BotCL with contrastive loss (BotCL$_\text{Cont}$) achieves the best accuracy on CUB200, ImageNet, and Synthetic, outperforming the baseline linear classifiers. It is also comparable to the state-of-the-art on MNIST and Synthetic. BotCL with reconstruction loss (BotCL$_\text{Rec}$) shows a performance drop over CUB200, ImageNet, and Synthetic, while it outperforms BotCL$_\text{Cont}$ on MNIST. This behavior supports our assumption that the reconstruction loss is useful only when concepts are strongly tied to their spatial position. Otherwise, $t$ is insufficient to reconstruct the original image, and BotCL fails. Contrastive self-supervision is the key to facilitating concept discovery. 

\begin{figure*}[!t]
\centering
\includegraphics[width=0.98\textwidth]{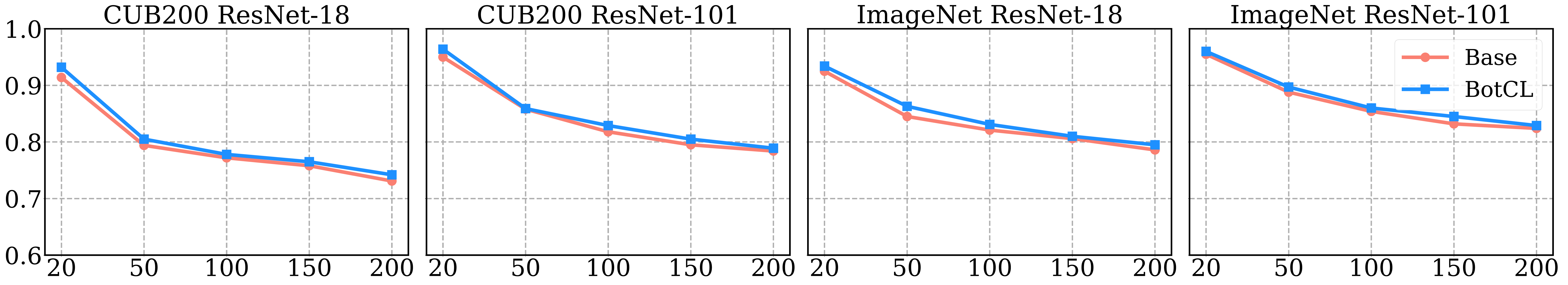}
\caption{Classification accuracy vs.~the number of classes. We used subsets of CUB200 and ImageNet with $k = 50$ and ResNet-18 and ResNet-101 backbones.}
\label{fig:acc_compare}
\end{figure*}

We also explore the relationship between the number of classes and BotCL's accuracy over CUB200 and ImageNet. We used small and large variants of ResNet as the backbone. We extracted subsets of the datasets consisting of the first $n$ classes along with the class IDs. \Cref{fig:acc_compare} shows that BotCL has a competitive performance when the number of classes is less than 200. We conclude that BotCL hardly degrades the classification performance on small- or middle-sized datasets. However, this is not the case for $n>200$ (refer to \textbf{supp.~material} for larger $n$ and different $k$'s).

\begin{table}[t]
\caption{Performance comparison in classification accuracy. The best concept-based method is highlighted in bold. BotCL$_\text{Rec}$ and BotCL$_\text{Cont}$ are both BotCL but with reconstruction and contrastive loss, respectively. For ImageNet, we used the first 200 classes.}
\centering
\resizebox{0.98\columnwidth}{!}{
\begin{tabular}{lcccc}
\toprule
  & CUB200 & ImageNet & MNIST & Synthetic\\
\midrule
\rowcolor{LightGrey}
Baseline & 0.731  & 0.786 & 0.988 & 0.999\\ \midrule
k-means$^*$ \cite{NEURIPS2020_ecb287ff} & 0.063 & 0.427 & 0.781 & 0.747\\
PCA$^*$ \cite{NEURIPS2020_ecb287ff} & 0.044  & 0.139 & 0.653 & 0.645\\ \midrule
SENN \cite{alvarez2018towards} & 0.642 & 0.673 & \textbf{0.985} & 0.984 \\
ProtoPNet \cite{chen2019looks} & 0.725 & 0.752 & 0.981 & 0.992 \\
BotCL$_\text{Rec}$ & 0.693  & 0.720 & 0.983 & 0.785\\
BotCL$_\text{Cont}$ & \textbf{0.740} & \textbf{0.795} & 0.980 & \textbf{0.998}\\
\bottomrule
\end{tabular}
}
\label{acc_tab}
\end{table}

\subsection{Interpretability}\label{lijie}
\subsubsection{Qualitative validation of discovered concepts}
\label{validity} 

\Cref{fig:mn-a} visualizes $a_\kappa$, showing concept $\kappa$ in the image, over MNIST. We selected 5 concepts out of 20 that are most frequently activated (\ie, $t_\kappa > 0.5$) in the training set.\footnote{\texttt{\small Cpts.1-5} are ordered based on the frequency counted in the dataset.} Taking digits \texttt{0} and \texttt{9} as an example, we can observe that they share \texttt{\small Cpts.3-5} and the only difference is \texttt{\small Cpt.2} that locates in the lower edge of the vertical stroke of \texttt{9}. This stroke is specific to digit \texttt{9}. We used BotCL$_\text{Rec}$, so we can remove \texttt{\small Cpt.2} before reconstruction, which generates an image like 0 (refer to \Cref{inference}). Some concepts are incompatible with human intuition; yet we can interpret such concepts (\eg, \texttt{\small Cpt.1} may attend to the missing stroke that completes a circle). 

For CUB200, we train BotCL$_\text{Const}$ with $n=50$ and $k=20$. \Cref{fig:bd-a} shows the attention maps of an image of \texttt{\small yellow headed black bird}. We can observe that the attentions for \texttt{\small Cpts.1-5} cover different body parts, including the head, wing, back, and feet, which proves that BotCL can learn valid concepts from the natural image as well. \textbf{Supp. material} exemplifies all concepts discovered from MNIST and CUB200.

\subsubsection{Consistency and distinctiveness of each concept}\label{top_activation} 
BotCL is designed to discover individually consistent and mutually distinctive concepts. We qualitatively verify this by showing each concept with its top-5 activated images\footnote{For each concept $\kappa$, five images whose $t_\kappa$ is highest among $\mathcal{D}$.} with attention maps in \Cref{fig:mn-b}. For MNIST, different concepts cover different patterns, and each concept covers the same patterns in different samples (even the samples of different classes). \Cref{fig:bd-b} for CUB200 shows that BotCL renders a similar behavior on the CUB200 dataset; that is, the top-5 concepts are responsible for different patterns, and each of them is consistent.

\subsubsection{Contribution of each concept in inference}\label{inference}
\begin{figure*}[!t]
  \begin{subfigure}{0pt}\phantomsubcaption\label{fig:mn-a}\end{subfigure}
  \begin{subfigure}{0pt}\phantomsubcaption\label{fig:mn-b}\end{subfigure}
  \begin{subfigure}{0pt}\phantomsubcaption\label{fig:mn-c}\end{subfigure}
\centering
\includegraphics[width=0.95\textwidth]{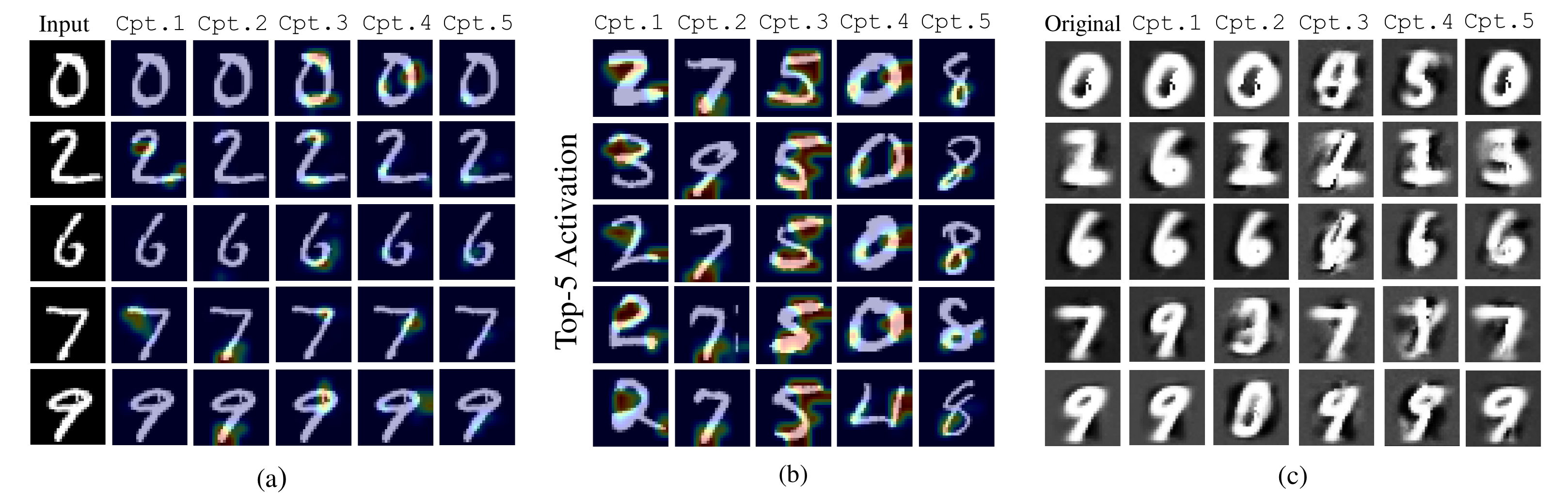}
\caption{Concepts for MNIST. (a) Attention maps for different input images. (b) Top-5 activated images (images in the dataset whose $t_\kappa$ is largest) for each concept. (c) Images reconstructed by our concept decoder with all detected concepts (original) and with a certain concept deactivated.}
\label{fig:mnist_demo}
\end{figure*}

We can qualitatively see the contribution of each concept by removing the concept and seeing the changes in the corresponding self-supervision task's output. As shown in \Cref{fig:mn-c}, when we set the activation of \texttt{\small Cpt.2} (responsible for the vertical stroke of digit \texttt{9}) to zero, the reconstructed image looks like digit \texttt{\small 0}. When \texttt{\small Cpt.1}, representing the absence of the circle in digit \texttt{7}, is deactivated (\ie, $t_1$ is set to 0), a circle emerges in the upper part of the reconstructed image. The resulting image looks more like digit \texttt{9}.

\begin{figure*}[!t]
  \begin{subfigure}{0pt}\phantomsubcaption\label{fig:bd-a}\end{subfigure}
  \begin{subfigure}{0pt}\phantomsubcaption\label{fig:bd-b}\end{subfigure}
  \begin{subfigure}{0pt}\phantomsubcaption\label{fig:bd-c}\end{subfigure}
\centering
\includegraphics[width=0.95\textwidth]{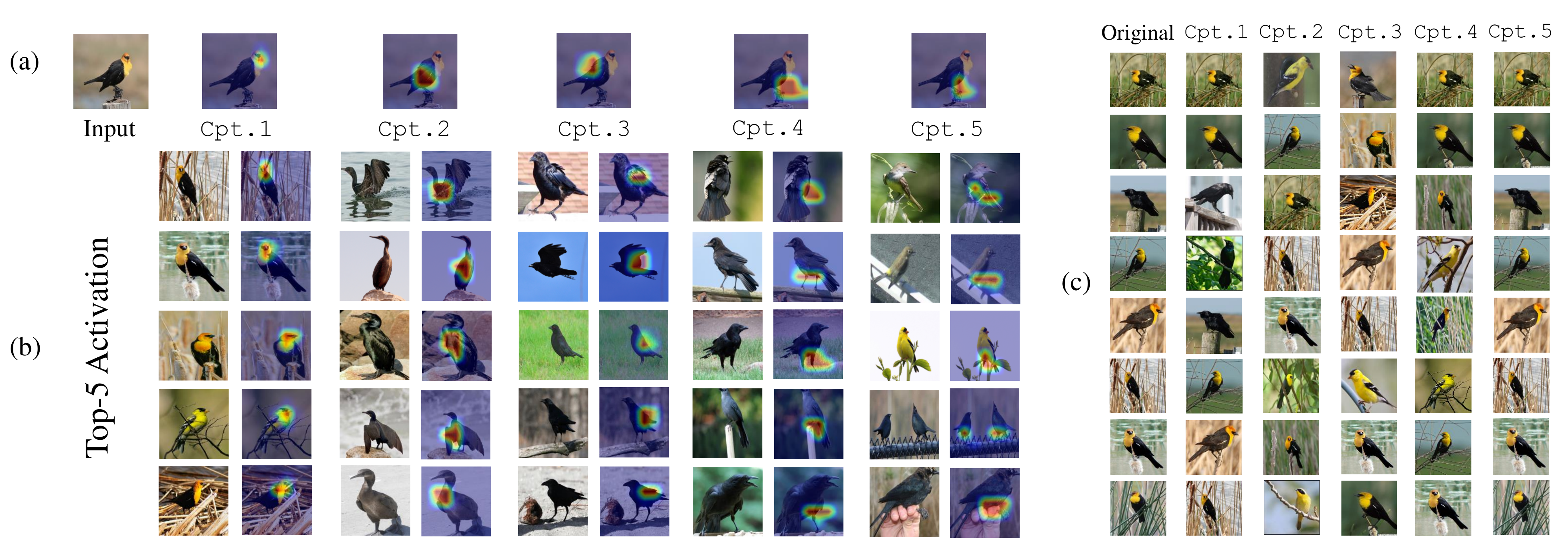}
\caption{Concepts learned for CUB200. (a) Visualization of 5 most important concepts for \texttt{\small yellow headed black bird}. (b) Top-5 activated concepts. (c) Image retrieval when all detected concepts were used (original) and when a certain concept was deactivated.}
\label{fig:bird}
\end{figure*}

For CUB200 shown in \Cref{fig:bd-c}, we show images most similar to the input image in \Cref{fig:bd-a} among the dataset in terms of $\hat{t}^\top \hat{t}'$, with ablating each concept. When \texttt{\small Cpt.1} (responsible for the yellow head) is deactivated, more black-head bird images appear in the top-8 images. \texttt{\small Cpt.5} covers birds' feet and is common among most bird classes. Deactivating this concept does not change the top-8 images. These observations suggest that although some concepts do not contribute to classification performance, images are successfully represented by combinations of concepts. 

\subsection{Quantitative Evaluation on Synthetic}\label{quan}

One problem of the concept-based approach is the absence of established quantitative evaluations of concepts because the choice of concepts may be arbitrary and the same level of representability may be achieved with different sets of concepts. A single predefined set of concepts is not enough to evaluate the goodness of discovered concepts. Literature has evaluated concepts qualitatively (as \Cref{lijie}) or by user study (as \Cref{sec:user_study}).

\begin{figure*}[!t]
  \begin{subfigure}{0pt}\phantomsubcaption\label{fig:sp-a}\end{subfigure}
  \begin{subfigure}{0pt}\phantomsubcaption\label{fig:sp-b}\end{subfigure}
\centering
\includegraphics[width=0.95\textwidth]{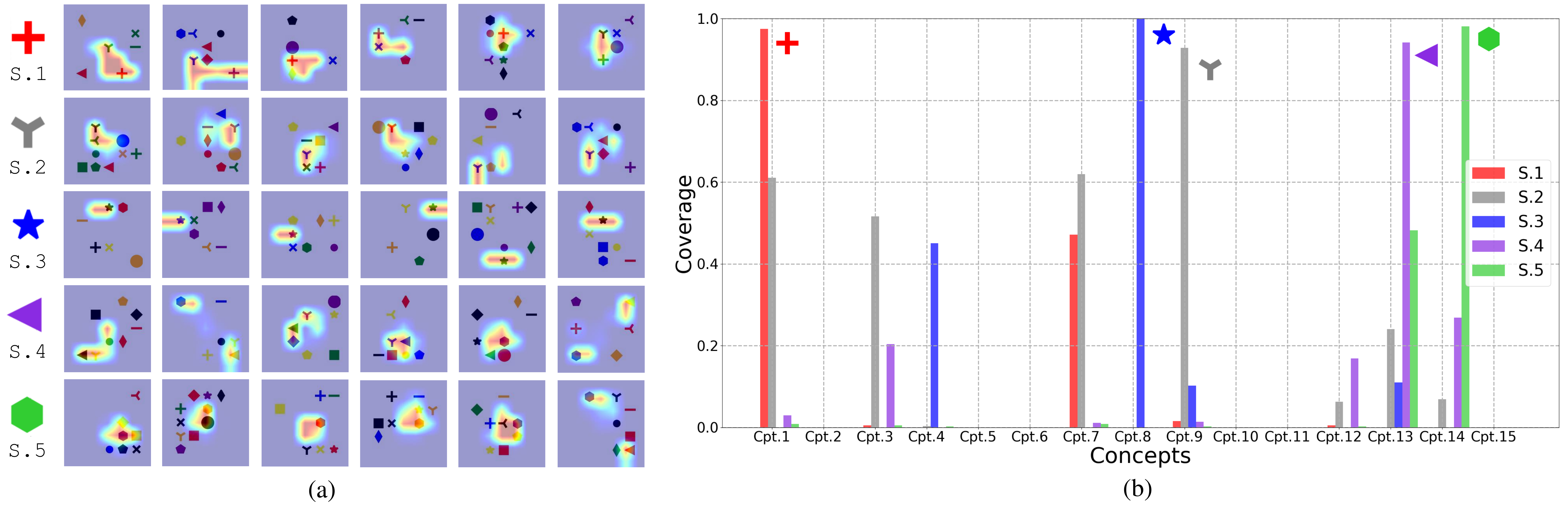}
\caption{Experiment on Synthetic. (a) \texttt{\small S.1}-\texttt{\small S.5} are the five shapes of which combinations form classes. Attention maps next to each shape are of the concept that covers the shape. (b) $\text{Coverage}_{s\kappa}$ (the concept associated with each of the five shapes is marked).}
\label{fig:shape}
\vspace{-0.1in}
\end{figure*}

We decided to use Synthetic \cite{NEURIPS2020_ecb287ff} for quantitatively evaluating concepts. The task is a multi-label classification that involves 15 shapes. Combinations of the 5 shapes (shown in \Cref{fig:sp-a}, \texttt{S.1} to \texttt{S.5}) form 15 classes, and the other 10 shapes are noises\footnote{Images are generated with random shapes, so there can be multiple classes (combinations of shapes) in a single image, which forms a multi-label classification task.}. We deem a shape is covered by concept $\kappa$ when the shape's area and concept $\kappa$'s area (the area with $a_\kappa > \gamma$ for BotCL, where $\gamma = 0.9$ is a predefined threshold) overlap. Let $h_{s\kappa} = 1$ denote shape $s$ overlaps with concept $\kappa$, and $h_{s\kappa} = 0$ otherwise. The coverage of $s$ by concept $\kappa$ is given by 
\begin{equation}
    \text{Coverage}_{s\kappa} = \mathbb{E}[h_{s\kappa}],
\end{equation}
where the expectation is computed over the images in the test set with concept $\kappa$ activated. The concepts and the 5 shapes are associated as a combinatorial optimization problem so that the sum of $\text{Coverage}_{s\kappa}$ over $s$ are maximized. 

We use $k=15$ to train BotCL. Figure \ref{fig:sp-a} visualizes the concept associated with each shape\footnote{Note that in this experiment, only shapes matter but not colors.}. A concept is located by $a_\kappa$ for 6 images with the highest concept activations $t_\kappa$. The concepts cover the associated shapes with relatively small regions, but one concept usually covers multiple shapes. This can be further evident in \Cref{fig:sp-b} that shows $\text{Coverage}_{s\kappa}$. \texttt{\small Cpt.8} only covers \texttt{S.3}, whereas \texttt{\small Cpt.1} and \texttt{\small Cpt.13} covers multiple shapes.  

We use three metrics other than accuracy to evaluate the performance of concept discovery\footnote{Further details are in \textbf{supp.~material}.}: (i) \textbf{Completeness} measures how well a concept covers its associated shape in the dataset. (ii) \textbf{Purity} shows the ability to discover concepts that only cover a single shape. (iii) \textbf{Distinctiveness} quantifies the difference among concepts based on the coverage. 

BotCL with contrastive loss is compared\footnote{SENN \cite{alvarez2018towards} and ProtoPNet \cite{chen2019looks} are not comparable. SENN's concepts globally cover a whole image. ProtoPNet requires way more concepts.}  with ACE \cite{ghorbani2019towards}, and two baselines PCA and k-means in \cite{NEURIPS2020_ecb287ff}. We apply k-means or PCA to $F$ of all images in the dataset after flattening the spatial dimensions. The cluster centers or the principal components are deemed as concepts. Attention maps can be computed by Euclidean distance or cosine similarity. Once the attention maps are obtained, we follow BotCL's process for classification. 

As shown in \Cref{quan_tab}, BotCL shows better completeness, distinctiveness, and accuracy scores than comparative methods. Although k-means is able to discover concepts, they are not optimized for the target classification task, and the performance is low. As we discussed, the concepts learned by BotCL tend to cover more than one target shape, causing a comparatively low purity. The cluster center of k-means is able to capture only one kind of shape at the cost of completeness. We can also observe that all methods are affected by concept number $k$, and generally a larger $k$ ensure better performance on all metrics. This result is not surprising, but we confirmed that a larger $k$ is preferable for better interpretability.  We detail the generation of the dataset, implementation of PCA and k-means, and formal definitions of metrics in \textbf{supp. material}.

\begin{table}[t]
\caption{Quantitative evaluation on Synthetic. Note that ACE uses concepts for post-hoc explanation and does not use them for classification. Comp., Dist., and Acc.~mean completeness, distinctiveness, and accuracy, respectively.}
\centering
\resizebox{0.8\columnwidth}{!}{
\begin{tabular}{llcccc}
\toprule
 & & Comp.  & Purity & Dist. & Acc. \\
\midrule
$k = 5$ & ACE  & 0.662 & 0.274 & 0.084 & ---  \\
 & k-means & 0.630 & 0.724 & 0.215 & 0.652  \\
 & PCA  & 0.458 & 0.170  & 0.298 & 0.571\\
 & BotCL  & 0.618 & 0.453 & 0.281 & 0.835  \\ \midrule
$k = 15$ & ACE & 0.614 & 0.221 & 0.151 & ---  \\
 & k-means & 0.816 & \textbf{0.978} & 0.272 & 0.747  \\
 & PCA  & 0.432 & 0.162 & 0.286 & 0.645  \\
 & BotCL  & \textbf{0.925} & 0.744 & \textbf{0.452} & \textbf{0.998}  \\
\bottomrule
\end{tabular}
}
\label{quan_tab}
\end{table}

\begin{table}[t]
\caption{Results of our user study.}
\centering
\resizebox{1.0\columnwidth}{!}{
\begin{tabular}{llcccccccc}
\toprule
 & &\multicolumn{2}{c}{CDR $\uparrow$} &\multicolumn{2}{c}{CC $\uparrow$} &\multicolumn{2}{c}{MIC $\downarrow$}\\
\cmidrule(lr){3-4} \cmidrule(lr){5-6} \cmidrule(lr){7-8}
Dataset & Concepts & Mean  & Std & Mean & Std & Mean & Std \\ 
\midrule
MNIST & Annotated & 1.000 & 0.000 & 0.838  & 0.150 & 0.071 & 0.047\\
 & BotCL & 0.825 & 0.288 & 0.581  & 0.274  & 0.199 & 0.072\\ 
 & Random & 0.122 & 0.070 & 0.163  & 0.074 & 0.438 & 0.039\\
\bottomrule
CUB200 & Annotated & 0.949 & 0.115 & 0.595  & 0.113 & 0.512 & 0.034\\
 & BotCL & 0.874 & 0.156 & 0.530  & 0.116 & 0.549 & 0.036\\ 
 & Random & 0.212 & 0.081 & 0.198  & 0.039 & 0.574 & 0.031\\
\bottomrule
\end{tabular}}
\label{user_}
\vspace{-0.1in}
\end{table}

\begin{figure*}[!t]
\centering
\begin{subfigure}{1.0\textwidth}
    \includegraphics[width=1.0\textwidth]{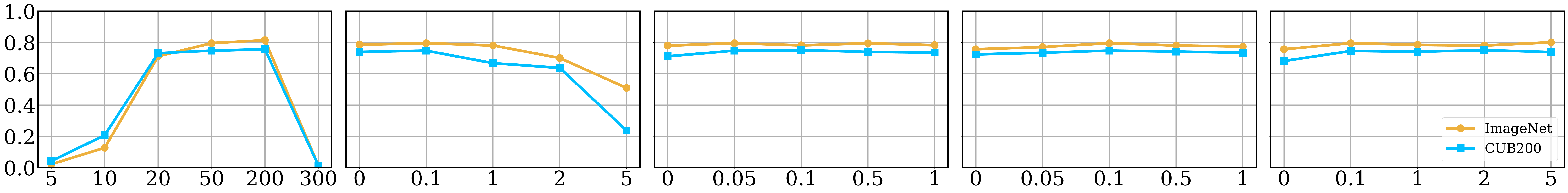}
    \caption{}
    \label{fig:ab-a}
  \end{subfigure}
  \hfill
  \begin{subfigure}{1.0\textwidth}
    \includegraphics[width=1.0\textwidth]{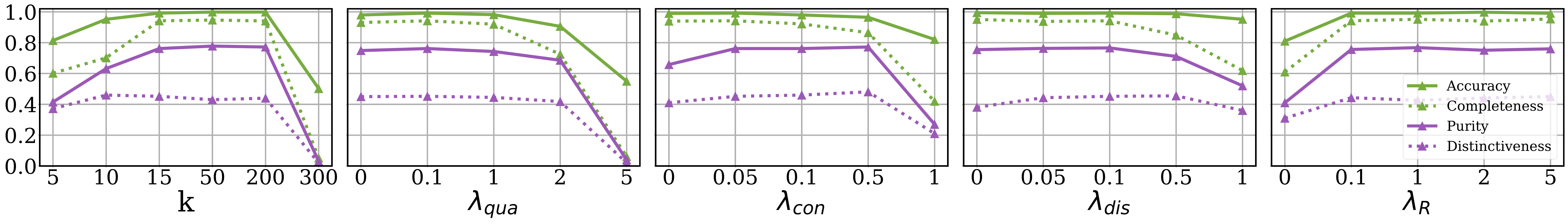}
    \caption{}
    \label{fig:ab-b}
  \end{subfigure}
\caption{Results of ablation study. (a) Hyperparameter values vs.~classification accuracy on ImageNet and CUB200. (b) Hyperparameter values vs.~classification accuracy and other metrics on  Synthetic.}
\label{fig:short}
\vspace{-0.05in}
\end{figure*}

\subsection{User Study}\label{sec:user_study}

Our user study is designed to evaluate BotCL with realistic datasets for the challenge of human understanding. Participants are asked to observe the test images with the attention map for concept $\kappa$ (refer to Section \ref{top_activation}) and select some phrases in the predefined vocabulary that best describes the concept (\ie, attended regions). They can also choose \textit{None of them} if they cannot find any consistent visual elements. 
We recruited 20 participants for each concept of MNIST and 30 participants for CUB200 using Amazon Mechanical Turk.

We defined three metrics to summarize the participants' responses. (i) \textbf{Concept discovery rate} (CDR): The ratio of the responses that are not \textit{None of them} to all responses. A higher CDR means participants can find some consistent visual elements for many concepts. (ii) \textbf{Concept consistency} (CC): The ratio of exact matches out of all pairs of participants' responses. A high value means many participants use the same phrases to describe a concept. (iii) \textbf{Mutual information between concepts} (MIC): The similarity of the response distribution, computed over all possible pairs of concepts. This value is high when multiple concepts cover the same visual elements; therefore, lower is better.

For comparison, we also evaluated a manual annotation\footnote{The authors annotated.} and random scribbling for the same images. \Cref{user_} shows that BotCL yields good scores for all metrics on both datasets (close to the manual annotation), showing the learned concepts are interpretable for humans (from CDR), consistent (from CC), and mutually distinct (from MIC). More details are in \textbf{supp. material}.

\subsection{Ablation Study}\label{Ablation}
We conducted ablation studies using the default hyperparameters except for the one to be explored. As there is no ground truth concept for CUB200 and ImageNet, only accuracy is evaluated (Figure \ref{fig:ab-a}). For Synthetic, accuracy and the three metrics in \Cref{quan} are employed ( \Cref{fig:ab-b}). 

\textbf{Impact of $k$.} A small $k$ decreases the accuracy and other metrics, which means the necessity of searching the minimum number of concepts. Also, training tends to fail for all datasets when $k$ is large (detailed in the \textbf{supp. materials}).  The number of concepts should be tuned for each dataset. This sensitivity is one of BotCL's limitations.

\textbf{Impact of $\lambda_{qua}$.} 
This hyperparameter controls how close $t$ should be to a binary. 
The accuracy and the other metrics worsen when $\lambda_{qua}$ gradually increases. BotCL encodes some information into $t$ (such as the area that a concept occupies), which is lost for larger $\lambda_{qua}$. An extreme value may also cause vanishing gradients. 

\textbf{Impact of $\lambda_{con}$ and $\lambda_{dis}$.} The individual consistency and mutual distinctiveness losses hardly affect the performance on CUB200 and ImageNet, although we can see a slight drop when the values are zero for CUB200. 
For Synthetic, the performance metrics vary as they are designed to be. Meanwhile, the accuracy is relatively insensitive to these hyperparameters. The choice of concepts may be highly arbitrary, and different sets of concepts may achieve similar classification performance. This arbitrariness may allow the designing of dedicated concept regularizers for the target task. However, 
training failures happen when they are set to be large. A small value benefits training.

\textbf{Impact of $\lambda_{R}$.} 
Due to the lower performance of the reconstruction loss, we studied the impact of the contrastive loss only.
The contrastive loss almost always improves the classification accuracy. The performance boost is significant in CUB200 and Synthetic. As ImageNet has more training data, this may imply that self-supervision greatly contributes to the learning of concepts when training samples are limited. These results demonstrate the importance of the contrastive loss. This is interesting since this loss uses the same labels as the classification loss.

\section{Conclusion}
This paper presents BotCL for learning bottleneck concepts. Our qualitative and quantitative evaluation showed BotCL's ability to learn concepts without explicit supervision on them but through training for a target classification task. We also demonstrated that BotCL could provide interpretability on its decision and learned concepts themselves.

\noindent\textbf{Limitations.} One limitation of BotCL is that it requires tuning the number $k$ of concepts for each dataset. It might be an interesting research direction to estimate $k$, \eg, based on the number of classes in a given classification task. We will investigate the phenomenon to mitigate this problem.

\paragraph{Acknowledgement} This work is partly supported by JST CREST Grant No. JPMJCR20D3, JST FOREST Grant No. JPMJFR216O, JSPS KAKENHI Grant-in-Aid for Scientific Research (A). This work is also supported by JSPS KAKENHI Grant Number 19K10662, 20K23343, 21K17764, and 22H03353.

{\small
\bibliographystyle{ieee_fullname}
\bibliography{egbib}

\begin{thebibliography}{10}\itemsep=-1pt

\bibitem{alvarez2018towards}
David Alvarez-Melis and Tommi~S Jaakkola.
\newblock Towards robust interpretability with self-explaining neural networks.
\newblock {\em NeurIPS}, 2018.

\bibitem{arrieta2020explainable}
Alejandro~Barredo Arrieta, Natalia D{\'\i}az-Rodr{\'\i}guez, Javier Del~Ser,
  Adrien Bennetot, Siham Tabik, Alberto Barbado, Salvador Garc{\'\i}a, Sergio
  Gil-L{\'o}pez, Daniel Molina, and Richard Benjamins.
\newblock Explainable artificial intelligence ({XAI}): Concepts, taxonomies,
  opportunities and challenges toward responsible {AI}.
\newblock {\em Information Fusion}, 58:82--115, 2020.

\bibitem{bach2015pixel}
Sebastian Bach, Alexander Binder, Gr{\'e}goire Montavon, Frederick Klauschen,
  Klaus-Robert M{\"u}ller, and Wojciech Samek.
\newblock On pixel-wise explanations for non-linear classifier decisions by
  layer-wise relevance propagation.
\newblock {\em PloS one}, 10(7):e0130140, 2015.

\bibitem{bau2017network}
David Bau, Bolei Zhou, Aditya Khosla, Aude Oliva, and Antonio Torralba.
\newblock Network dissection: Quantifying interpretability of deep visual
  representations.
\newblock In {\em CVPR}, pages 6541--6549, 2017.

\bibitem{bucher2018semantic}
Maxime Bucher, St{\'e}phane Herbin, and Fr{\'e}d{\'e}ric Jurie.
\newblock Semantic bottleneck for computer vision tasks.
\newblock In {\em ACCV}, pages 695--712, 2018.

\bibitem{chattopadhay2018grad}
Aditya Chattopadhay, Anirban Sarkar, Prantik Howlader, and Vineeth~N
  Balasubramanian.
\newblock Grad-{CAM}++: {G}eneralized gradient-based visual explanations for
  deep convolutional networks.
\newblock In {\em WACV}, pages 839--847, 2018.

\bibitem{chefer2021transformer}
Hila Chefer, Shir Gur, and Lior Wolf.
\newblock Transformer interpretability beyond attention visualization.
\newblock In {\em CVPR}, pages 782--791, 2021.

\bibitem{chen2019looks}
Chaofan Chen, Oscar Li, Daniel Tao, Alina Barnett, Cynthia Rudin, and
  Jonathan~K Su.
\newblock This looks like that: Deep learning for interpretable image
  recognition.
\newblock {\em NeurIPS}, 32, 2019.

\bibitem{chen2020simple}
Ting Chen, Simon Kornblith, Mohammad Norouzi, and Geoffrey Hinton.
\newblock A simple framework for contrastive learning of visual
  representations.
\newblock In {\em ICML}, pages 1597--1607, 2020.

\bibitem{deng2009imagenet}
Jia Deng, Wei Dong, Richard Socher, Li-Jia Li, Kai Li, and Li Fei-Fei.
\newblock Image{N}et: {A} large-scale hierarchical image database.
\newblock In {\em CVPR}, pages 248--255, 2009.

\bibitem{deng2012mnist}
Li Deng.
\newblock The mnist database of handwritten digit images for machine learning
  research.
\newblock {\em Signal Processing Magazine}, 29(6):141--142, 2012.

\bibitem{fong2019understanding}
Ruth Fong, Mandela Patrick, and Andrea Vedaldi.
\newblock Understanding deep networks via extremal perturbations and smooth
  masks.
\newblock In {\em ICCV}, pages 2950--2958, 2019.

\bibitem{ge2021peek}
Yunhao Ge, Yao Xiao, Zhi Xu, Meng Zheng, Srikrishna Karanam, Terrence Chen,
  Laurent Itti, and Ziyan Wu.
\newblock A peek into the reasoning of neural networks: Interpreting with
  structural visual concepts.
\newblock In {\em CVPR}, pages 2195--2204, 2021.

\bibitem{ghorbani2019towards}
Amirata Ghorbani, James Wexler, James Zou, and Been Kim.
\newblock Towards automatic concept-based explanations.
\newblock {\em NeurIPS}, 2019.

\bibitem{he2022transfg}
Ju He, Jie-Neng Chen, Shuai Liu, Adam Kortylewski, Cheng Yang, Yutong Bai, and
  Changhu Wang.
\newblock Transfg: A transformer architecture for fine-grained recognition.
\newblock In {\em AAAI}, pages 852--860, 2022.

\bibitem{he2020momentum}
Kaiming He, Haoqi Fan, Yuxin Wu, Saining Xie, and Ross Girshick.
\newblock Momentum contrast for unsupervised visual representation learning.
\newblock In {\em CVPR}, pages 9729--9738, 2020.

\bibitem{he2016deep}
Kaiming He, Xiangyu Zhang, Shaoqing Ren, and Jian Sun.
\newblock Deep residual learning for image recognition.
\newblock In {\em CVPR}, pages 770--778, 2016.

\bibitem{hirota2022quantifying}
Yusuke Hirota, Yuta Nakashima, and Noa Garcia.
\newblock Quantifying societal bias amplification in image captioning.
\newblock In {\em CVPR}, pages 13450--13459, 2022.

\bibitem{holzinger2019causability}
Andreas Holzinger, Georg Langs, Helmut Denk, Kurt Zatloukal, and Heimo
  M{\"u}ller.
\newblock Causability and explainability of artificial intelligence in
  medicine.
\newblock {\em Wiley Interdisciplinary Reviews: Data Mining and Knowledge
  Discovery}, 9(4):e1312, 2019.

\bibitem{huang2020interpretable}
Zixuan Huang and Yin Li.
\newblock Interpretable and accurate fine-grained recognition via region
  grouping.
\newblock In {\em CVPR}, pages 8662--8672, 2020.

\bibitem{kim2018interpretability}
Been Kim, Martin Wattenberg, Justin Gilmer, Carrie Cai, James Wexler, Fernanda
  Viegas, and Rory Sayres.
\newblock Interpretability beyond feature attribution: Quantitative testing
  with concept activation vectors {TCAV}.
\newblock In {\em ICML}, pages 2668--2677, 2018.

\bibitem{koh2020concept}
Pang~Wei Koh, Thao Nguyen, Yew~Siang Tang, Stephen Mussmann, Emma Pierson, Been
  Kim, and Percy Liang.
\newblock Concept bottleneck models.
\newblock In {\em ICML}, pages 5338--5348, 2020.

\bibitem{kumar2009attribute}
Neeraj Kumar, Alexander~C Berg, Peter~N Belhumeur, and Shree~K Nayar.
\newblock Attribute and simile classifiers for face verification.
\newblock In {\em ICCV}, pages 365--372, 2009.

\bibitem{lake2015human}
Brenden~M Lake, Ruslan Salakhutdinov, and Joshua~B Tenenbaum.
\newblock Human-level concept learning through probabilistic program induction.
\newblock {\em Science}, 350(6266):1332--1338, 2015.

\bibitem{laugel2019dangers}
Thibault Laugel, Marie-Jeanne Lesot, Christophe Marsala, Xavier Renard, and
  Marcin Detyniecki.
\newblock The dangers of post-hoc interpretability: Unjustified counterfactual
  explanations.
\newblock {\em arXiv preprint arXiv:1907.09294}, 2019.

\bibitem{li2021scouter}
Liangzhi Li, Bowen Wang, Manisha Verma, Yuta Nakashima, Ryo Kawasaki, and
  Hajime Nagahara.
\newblock Scouter: Slot attention-based classifier for explainable image
  recognition.
\newblock In {\em ICCV}, pages 1046--1055, 2021.

\bibitem{locatello2020object}
Francesco Locatello, Dirk Weissenborn, Thomas Unterthiner, Aravindh Mahendran,
  Georg Heigold, Jakob Uszkoreit, Alexey Dosovitskiy, and Thomas Kipf.
\newblock Object-centric learning with slot attention.
\newblock {\em NeurIPS}, 2020.

\bibitem{losch2019interpretability}
Max Losch, Mario Fritz, and Bernt Schiele.
\newblock Interpretability beyond classification output: Semantic bottleneck
  networks.
\newblock {\em arXiv preprint arXiv:1907.10882}, 2019.

\bibitem{petsiuk2018rise}
Vitali Petsiuk, Abir Das, and Kate Saenko.
\newblock {RISE}: {R}andomized input sampling for explanation of black-box
  models.
\newblock {\em BMVC}, 2018.

\bibitem{posada2022eclad}
Andres~Felipe Posada-Moreno, Nikita Surya, and Sebastian Trimpe.
\newblock {ECLAD}: Extracting concepts with local aggregated descriptors.
\newblock {\em arXiv preprint arXiv:2206.04531}, 2022.

\bibitem{rigotti2022attention}
Mattia Rigotti, Christoph Miksovic, Ioana Giurgiu, Thomas Gschwind, and Paolo
  Scotton.
\newblock Attention-based interpretability with concept transformers.
\newblock In {\em ICLR}, 2022.

\bibitem{schulz2020restricting}
Karl Schulz, Leon Sixt, Federico Tombari, and Tim Landgraf.
\newblock Restricting the flow: Information bottlenecks for attribution.
\newblock {\em arXiv preprint arXiv:2001.00396}, 2020.

\bibitem{selvaraju2017grad}
Ramprasaath~R Selvaraju, Michael Cogswell, Abhishek Das, Ramakrishna Vedantam,
  Devi Parikh, and Dhruv Batra.
\newblock Grad-{CAM}: Visual explanations from deep networks via gradient-based
  localization.
\newblock In {\em CVPR}, pages 618--626, 2017.

\bibitem{shi2019knowledge}
Botian Shi, Lei Ji, Pan Lu, Zhendong Niu, and Nan Duan.
\newblock Knowledge aware semantic concept expansion for image-text matching.
\newblock In {\em IJCAI}, volume~1, page~2, 2019.

\bibitem{shrikumar2017learning}
Avanti Shrikumar, Peyton Greenside, and Anshul Kundaje.
\newblock Learning important features through propagating activation
  differences.
\newblock In {\em ICML}, pages 3145--3153, 2017.

\bibitem{simonyan2014deep}
Karen Simonyan, Andrea Vedaldi, and Andrew Zisserman.
\newblock Deep inside convolutional networks: Visualising image classification
  models and saliency maps.
\newblock In {\em ICLR Workshop}, 2014.

\bibitem{stammer2022interactive}
Wolfgang Stammer, Marius Memmel, Patrick Schramowski, and Kristian Kersting.
\newblock Interactive disentanglement: Learning concepts by interacting with
  their prototype representations.
\newblock In {\em CVPR}, pages 10317--10328, 2022.

\bibitem{van2022explainable}
Bas~HM van~der Velden, Hugo~J Kuijf, Kenneth~GA Gilhuijs, and Max~A Viergever.
\newblock Explainable artificial intelligence ({XAI}) in deep learning-based
  medical image analysis.
\newblock {\em Medical Image Analysis}, page 102470, 2022.

\bibitem{varshneya2021learning}
Saurabh Varshneya, Antoine Ledent, Robert~A Vandermeulen, Yunwen Lei, Matthias
  Enders, Damian Borth, and Marius Kloft.
\newblock Learning interpretable concept groups in {CNN}s.
\newblock {\em IJCAI}, 2021.

\bibitem{vaswani2017attention}
Ashish Vaswani, Noam Shazeer, Niki Parmar, Jakob Uszkoreit, Llion Jones,
  Aidan~N Gomez, {\L}ukasz Kaiser, and Illia Polosukhin.
\newblock Attention is all you need.
\newblock In {\em NeurIPS}, pages 5998--6008, 2017.

\bibitem{wang2021mtunet}
Bowen Wang, Liangzhi Li, Manisha Verma, Yuta Nakashima, Ryo Kawasaki, and
  Hajime Nagahara.
\newblock Mtunet: Few-shot image classification with visual explanations.
\newblock In {\em CVPR workshops}, pages 2294--2298, 2021.

\bibitem{wang2019designing}
Danding Wang, Qian Yang, Ashraf Abdul, and Brian~Y Lim.
\newblock Designing theory-driven user-centric explainable {AI}.
\newblock In {\em Proc. CHI conference on human factors in computing systems},
  pages 1--15, 2019.

\bibitem{wang2020score}
Haofan Wang, Zifan Wang, Mengnan Du, Fan Yang, Zijian Zhang, Sirui Ding, Piotr
  Mardziel, and Xia Hu.
\newblock Score-{CAM}: Score-weighted visual explanations for convolutional
  neural networks.
\newblock In {\em CVPR workshops}, pages 24--25, 2020.

\bibitem{welinder2010caltech}
Peter Welinder, Steve Branson, Takeshi Mita, Catherine Wah, Florian Schroff,
  Serge Belongie, and Pietro Perona.
\newblock Caltech-{UCSD} birds 200.
\newblock 2010.

\bibitem{context}
Liang~Wang Yan~Huang, Qi~Wu.
\newblock Learning semantic concepts and order for image and sentence matching.
\newblock In {\em CVPR}, 2018.

\bibitem{NEURIPS2020_ecb287ff}
Chih-Kuan Yeh, Been Kim, Sercan Arik, Chun-Liang Li, Tomas Pfister, and Pradeep
  Ravikumar.
\newblock On completeness-aware concept-based explanations in deep neural
  networks.
\newblock In {\em NeurIPS}, volume~33, pages 20554--20565, 2020.

\bibitem{zhang2018unreasonable}
Richard Zhang, Phillip Isola, Alexei~A Efros, Eli Shechtman, and Oliver Wang.
\newblock The unreasonable effectiveness of deep features as a perceptual
  metric.
\newblock In {\em CVPR}, pages 586--595, 2018.

\bibitem{zhou2014object}
Bolei Zhou, Aditya Khosla, Agata Lapedriza, Aude Oliva, and Antonio Torralba.
\newblock Object detectors emerge in deep scene {CNN}s.
\newblock {\em ICLR}, 2015.

\bibitem{zhou2016learning}
Bolei Zhou, Aditya Khosla, Agata Lapedriza, Aude Oliva, and Antonio Torralba.
\newblock Learning deep features for discriminative localization.
\newblock In {\em CVPR}, pages 2921--2929, 2016.

\bibitem{zhou2018interpretable}
Bolei Zhou, Yiyou Sun, David Bau, and Antonio Torralba.
\newblock Interpretable basis decomposition for visual explanation.
\newblock In {\em ECCV}, pages 119--134, 2018.

\end{thebibliography}
}

\end{document}